\documentclass[journal]{IEEEtran}
\usepackage{amsmath,amsfonts}
\usepackage{array}
\usepackage{textcomp}
\usepackage{stfloats}
\usepackage{url}
\usepackage{verbatim}
\usepackage{graphicx}
\usepackage{cite}
\usepackage{tikz}
\usepackage{comment}
\usepackage{amsmath,amssymb} 
\usepackage{color}
\usepackage{graphicx}
\usepackage{caption}
\usepackage{subcaption}
\usepackage{booktabs}
\usepackage{multirow}
\usepackage[ruled, linesnumbered]{algorithm2e}
\usepackage{algpseudocode}
\usepackage{bbding}
\usepackage{wrapfig}
\usepackage{picinpar}
\SetKwProg{Init}{Init}{}{}
\SetKw{And}{and}

\usepackage[colorlinks,
            linkcolor=red,
            anchorcolor=blue,
            citecolor=green
            ]{hyperref}

\usepackage[capitalize]{cleveref}
\crefname{section}{Sec.}{Secs.}
\Crefname{section}{Section}{Sections}
\Crefname{table}{Table}{Tables}
\crefname{table}{Tab.}{Tabs.}

\usepackage{xspace}
\newcommand{\name}[0]{GMConv\xspace}

\usepackage{xcolor}

\makeatletter
\DeclareRobustCommand\onedot{\futurelet\@let@token\@onedot}
\def\@onedot{\ifx\@let@token.\else.\null\fi\xspace}
\def\eg{\emph{e.g}\onedot}

\def\etal{\emph{et al}\onedot}

\usepackage{scalerel}
\usepackage{tikz}
\usetikzlibrary{svg.path}
\definecolor{orcidlogocol}{HTML}{A6CE39}
\tikzset{
    orcidlogo/.pic={
        \fill[orcidlogocol] svg{M256,128c0,70.7-57.3,128-128,128C57.3,256,0,198.7,0,128C0,57.3,57.3,0,128,0C198.7,0,256,57.3,256,128z};
        \fill[white] svg{M86.3,186.2H70.9V79.1h15.4v48.4V186.2z}
        svg{M108.9,79.1h41.6c39.6,0,57,28.3,57,53.6c0,27.5-21.5,53.6-56.8,53.6h-41.8V79.1z M124.3,172.4h24.5c34.9,0,42.9-26.5,42.9-39.7c0-21.5-13.7-39.7-43.7-39.7h-23.7V172.4z}
        svg{M88.7,56.8c0,5.5-4.5,10.1-10.1,10.1c-5.6,0-10.1-4.6-10.1-10.1c0-5.6,4.5-10.1,10.1-10.1C84.2,46.7,88.7,51.3,88.7,56.8z};
    }
}

\newcommand\orcidicon[1]{\href{https://orcid.org/#1}{\mbox{\scalerel*{
                \begin{tikzpicture}[yscale=-1,transform shape]
                \pic{orcidlogo};
                \end{tikzpicture}
            }{|}}}}

\begin{document}

\title{GMConv: Modulating Effective Receptive Fields for Convolutional Kernels}

\author{~Qi~Chen, Chao Li, Jia Ning, Stephen Lin,~\IEEEmembership{Member,~IEEE}, ~and~Kun~He$^\dag$ ${\textsuperscript{\orcidicon{0000-0001-7627-4604}}}$,~\IEEEmembership{Senior~Member,~IEEE}
\IEEEcompsocitemizethanks{\IEEEcompsocthanksitem Qi Chen, Chao Li, Jia Ning and Kun He are with Huazhong University of Science and Technology, Hubei, China. 
Stephen Lin is with Microsoft Research Asia. 
E-mail: \{bloom24,~D201880880,~ninja\}@hust.edu.cn,~stevelin@microsoft.com,~brooklet60@hust.edu.cn}
}

\maketitle



\begin{abstract}
In convolutional neural networks, the convolutions are conventionally performed using a square kernel with a fixed $N \times N$ receptive field (RF).
However, what matters most to the network is the effective receptive field (ERF) that indicates the extent with which input pixels contribute to an output pixel. 
Inspired by the property that ERFs typically exhibit a Gaussian distribution,
we propose a Gaussian Mask convolutional kernel (GMConv) in this work. Specifically, GMConv utilizes the Gaussian function to generate a concentric symmetry mask that is placed over the kernel to refine the RF. Our GMConv can directly replace the standard convolutions in existing CNNs and can be easily trained end-to-end by standard back-propagation. We evaluate our approach through extensive experiments on image classification and object detection tasks. Over several tasks and standard base models, our approach compares favorably against the standard convolution. For instance, using GMConv for AlexNet and ResNet-50, the top-1 accuracy on ImageNet classification is boosted by $0.98\%$ and $0.85\%$, respectively.


\end{abstract}


\section{INTRODUCTION}
\label{sec:intro}

Convolutional Neural Networks (CNNs) have significantly boosted the performance of computer vision tasks, including image classification~\cite{NIPS2012_c399862d,he2016deep}, object detection~\cite{girshick14CVPR,renNIPS15fasterrcnn}, and many other applications~\cite{SunXLW19,lu2020retinatrack}. Their great success is attributed to their ability to learn hierarchical representations of the input data, through the use of convolutional and pooling layers. Despite their remarkable achievements, the recent emergence of Vision Transformers (ViTs)~\cite{dosovitskiy2020vit,liu2021Swin,liu2021swinv2,wang2021pyramid} has attracted increasing attention, as they offer a promising alternative to CNNs and have shown remarkable results in various visual recognition tasks, outperforming CNNs in some cases. 

\begin{figure}[t]
  \centering
  \includegraphics[width=1\linewidth]{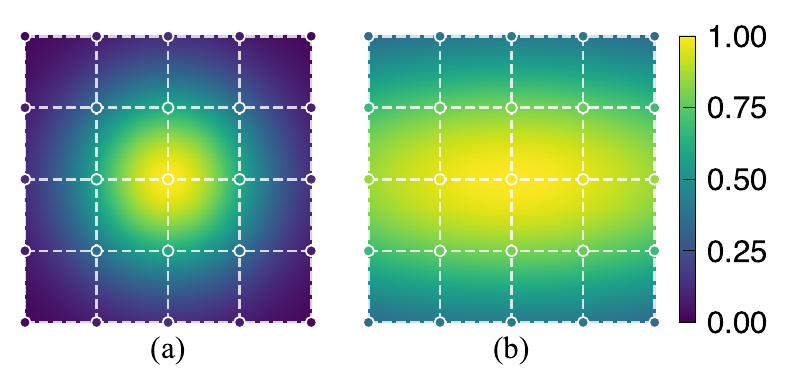}
  \caption{Typical receptive fields of $5 \times 5$ Gaussian mask convolutions.
  (a) Receptive field of \name with circular mask.
  (b) Receptive field of \name with elliptic mask.
  }
  \label{fig:f1}
\end{figure}

Meanwhile, there has been a continued effort to improve convolutional neural networks, with many recent works dedicated to the design of CNN architectures. For instance, the recently proposed large kernel CNNs~\cite{ding2022scaling,liu2022convnet,liu2022more} show that these pure CNN architectures are competitive with state-of-the-art ViTs in accuracy, scalability, and robustness across all major benchmarks. These efforts have renewed the attention on CNNs and have encouraged the community to reconsider the importance of convolutions for computer vision.
On the other hand, square convolutions have long been used as the standard structure for building various CNN architectures~\cite{szegedy2015going,he2016deep,xie2017aggregated,hu2018senet,radosavovic2020designing} as they are well-suited for the rectangular geometry of matrices and tensor computation.
Although there have been some works on designing new convolutional kernels by sampling pixels with offsets in recent years~\cite{jeon2017active,dai2017deformable,zhu2019deformable,He2021CirKernel}, these efforts are relatively few compared to the vast number of studies devoted to CNN architecture design.

The receptive field (RF) is a basic concept in CNNs, 
corresponding to the region in the input on which a CNN unit depends.
The research of Luo \etal~\cite{luo2016understanding} further introduces the concept of effective receptive field (ERF), indicating an effective area within the receptive field that primarily influences the response of an output unit. They emphasize that the ERF, which shows how much each pixel contributes to the output, is what really matters to the network. The ERF cannot be modeled by simply sampling pixels with offsets.
In this work, we aim to simulate the ERF from a different perspective.



Inspired by the fact that the ERF exhibits a Gaussian distribution, we propose a Gaussian Mask Convolutional Kernel (\name) to introduce a circular (or elliptic) concentric RF into the CNN kernels. It adds $\sigma$ ($\sigma_1$, $\sigma_2$) parameters in the standard convolution and generates an RF adjustment mask as a Gaussian-like function. 
Different from existing related works~\cite{jeon2017active,dai2017deformable,zhu2019deformable,HeCircularKernel2021} that focus on designing shapes of convolutional kernels, we 
focus on adding masks on the square kernels to change the convolution operation.
The effect of our \name is illustrated in Figure~\ref{fig:f1}.
The values on the mask determine the sensitivity of the convolutional kernel to different regions, with larger values indicating stronger sensitivity and smaller values indicating weaker sensitivity. 
As the parameters vary, the RFs of \name show a progressive circular (elliptical) concentric distribution, which could approximate effective receptive fields. 

Specifically, we introduce two kinds of \name. 
The first is the static version of \name (S-\name), which only requires one additional parameter ($\sigma$) compared to the standard convolution.
This parameter enables the convolution to generate a corresponding circular RF mask, allowing it to model spatial relationships and adjust the RF. 
The second is the dynamic version of \name (D-\name), which further strengthens \name with more parameters that control the mask distribution from both horizontal and vertical aspects, as well as a dynamic sigma module, which predicts specialized sigma parameters dynamically for each input. Both modules are lightweight and can be easily integrated into existing CNNs as a drop-in replacement for standard convolutions.


Our experiments demonstrate that replacing the standard convolutional layers with our proposed Gaussian Mask Convolutional Kernel (\name) can considerably boost the performance of equivalent neural networks on ImageNet classification and COCO object detection. Specifically, we found that the use of \name can greatly enhance the localization accuracy of small and medium-sized objects, which is a critical requirement for many computer vision applications.
Additionally, we also observe that \name has a tendency to shrink the receptive fields of convolutions in shallow layers while preserving the default receptive field size in deeper layers.

Our main contributions are as follows:
\begin{itemize}
\setlength{\itemsep}{0pt}
\setlength{\parsep}{0pt}
\setlength{\parskip}{0pt}
    \item We propose \name, a straightforward and effective convolution, that generates a Gaussian distribution-like mask to introduce concentric circular (elliptic) receptive fields in convolutional kernels. 
    \item We design both a static and a dynamic version of \name, which can be easily integrated with commonly used CNN architectures with negligible extra parameters and complexity.
    \item We conduct a series of comprehensive experiments to demonstrate that \name can improve the performance of various standard benchmark models across three benchmark datasets: CIFAR datasets (CIFAR-10 and CIFAR-100), ImageNet, and COCO 2017.
    \item Our 
    studies of \name suggest a potential preference for smaller receptive fields in shallower layers of CNNs and larger ones in deeper layers. These findings may offer insights for future convolutional neural networks design.
    
\end{itemize}

\section{RELATED WORK}
\label{sec:rw}
\subsection{CNN Structure Design}
Since AlexNet~\cite{NIPS2012_c399862d} revealed the potential of CNNs,  
there have emerged numerous powerful manually designed networks~\cite{simonyan2014very,he2016deep,xie2017aggregated,hu2018senet} and automatic architectures~\cite{tan2019efficientnet,radosavovic2020designing}.
These networks are typically composed of three distinct parts, namely the stem, body, and head~\cite{radosavovic2020designing}.
In general, the stem component is an $N \times N$ convolution with stride two that halves the height and width of an input image.
The head component is tailored towards a specific task, \eg, global average pooling followed by a fully connected layer for predicting the output class in the image classification task.
The body component consists of multiple stages of transformation that progressively reduce the spatial resolution of the feature maps. Generally, the stem and head are kept fixed and simple, and most works are focused on the structure of the network body, which is more critical to network accuracy.
Based on the design pattern, when applying \name into a CNN, we use different versions of \name for the stem layer and body layer. Details will be presented in~\cref{sec:method}.

\subsection{Convolutional Kernel Design}
Grouped convolutions~\cite{NIPS2012_c399862d} divide the feature maps to multiple GPUs to process in parallel and finally fuse the results. This approach has inspired the design of follow-up lightweight networks. Depthwise separable convolutions~\cite{chollet2017xception,howard2017mobilenets} consist of a depthwise convolution followed by a pointwise convolution. The depthwise convolution first performs a spatial grouped convolution independently over each channel of the input. Then the pointwise convolution applies a $1 \times 1$ convolution to fuse the channel outputs of the depthwise convolution. This approach achieves a good trade-off between accuracy and resource efficiency,
leading to broad use in lightweight computing devices.

The above variants inherit the square kernel in general, thus resulting in a fixed receptive field. 
To enable more flexible receptive fields and better learn the spatial transformation from the input, 
another line of studies~\cite{jeon2017active,dai2017deformable,zhu2019deformable} designs more free-form convolutions with deformations.
Active convolutions~\cite{jeon2017active} and deformable convolutions~\cite{dai2017deformable,zhu2019deformable} augment the sampling locations in the receptive fields with dynamic offsets that are learned from the preceding features via additional convolution layers.
Deformable ConvNets v2~\cite{zhu2019deformable} further introduces a dynamic modulation mechanism that evaluates the importance of each sampled location but with higher computational overhead. Deformable Kernel~\cite{DBLP:conf/iclr/GaoZLD20} resamples the original kernel space towards 
recovering the deformation of objects. 
In contrast, our focus is transforming the receptive field inside the original kernel. 
It gives convolutions more flexible receptive fields without changing the tensor computation for standard convolutions.

There are also some works using masked convolutions~\cite{ma2019macow,krull2019noise2void,jain2020locally}. MaConw~\cite{ma2019macow} and LMConv~\cite{jain2020locally} use binary masks to generate models. Noise2void~\cite{krull2019noise2void} employs a binary mask for denoising.
In contrast, \name applies a Gaussian-like mask to model the effective receptive field for the image classification task. 

\subsection{Dynamic Mechanism}
A dynamic mechanism has been introduced into models to elevate their power. Benefiting from input dependency, networks can adjust themselves to fit diverse inputs automatically and improve representation ability.
From the perspective of dynamic features,
SE-Net~\cite{hu2018senet} proposes the ``Squeeze-and-Excitation" block, which adaptively recalibrates channel-wise features by explicitly modeling the inter-dependencies among channels. The subsequently proposed CBAM block~\cite{woo2018cbam} recalibrates both channel-wise and spatial-wise features.
From the aspect of dynamic convolution kernels,
the deformable convolutions~\cite{jeon2017active,dai2017deformable,zhu2019deformable} learn offsets for each input to enable adaptive geometric variations.
CondConv~\cite{yang2019condconv} and Dynamic Convolution~\cite{chen2020dynamic} dynamically aggregate multiple parallel convolution kernels to increase the model capacity.
Different from the above methods, the dynamic form of \name, called D-\name, applies a dynamic RF mask over the kernel to enable adaptive RF on square convolutions.


\subsection{Gaussian-Distributed Effective Receptive Field} 
Luo \etal~\cite{luo2016understanding} find that the effective receptive field, which shows the main contributions to a unit's output, only occupies a small fraction of the actual RF and exhibits a Gaussian distribution. This finding has inspired works that incorporate the Gaussian distribution to convolutions. In image segmentation, the work of Shelhamer \etal~\cite{shelhamer2019efficient} and Sun \etal~\cite{sun2021gaussian} use a Gaussian distribution to construct deformable convolutions. They generally follow the line of learning offsets for geometric variations.





\section{METHODOLOGY}
\label{sec:method}

\begin{figure*}[!t]
\centering
     \begin{minipage}[b]{0.601\linewidth}
         \centering
         \includegraphics[width=\linewidth]{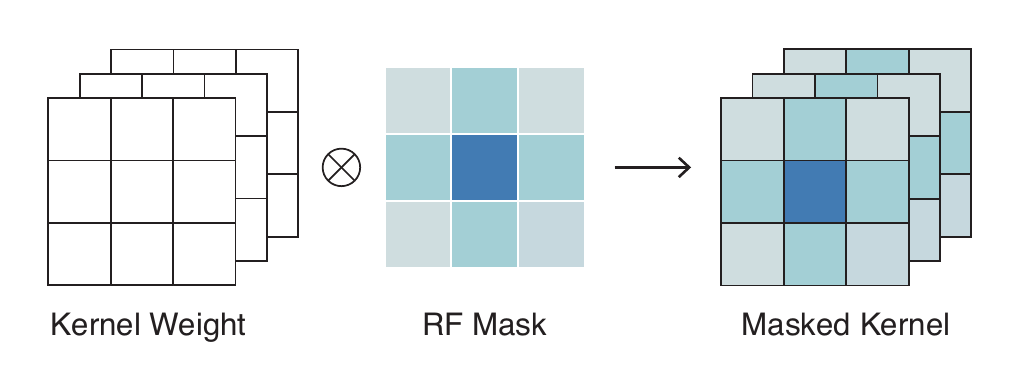}
         \caption{Overview of \name.}
         \label{fig:overview_of_gmc}
     \end{minipage}
     \hfill
     \begin{minipage}[b]{0.389\linewidth}
         \centering
         \includegraphics[width=\linewidth]{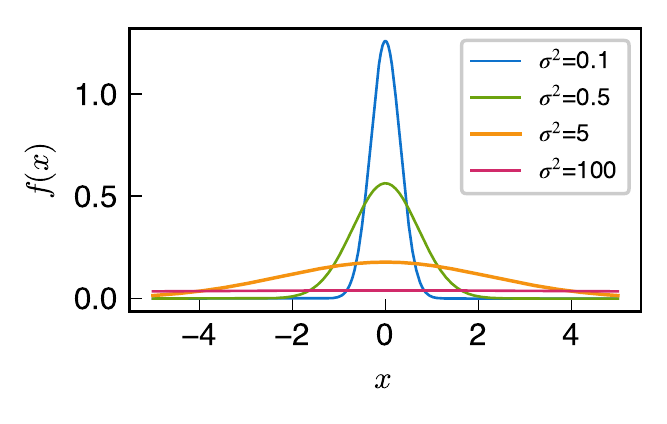}
         \caption{Curves of standard one-dimensional Gaussian function. All $\mu$ values are set to 0 for clearer comparison.}
         \label{fig:gauss_curve}
     \end{minipage}
\end{figure*}

In this section, we introduce the details of \name. First, we present an overview of \name. Then, we illustrate how to generate the RF masks. Next, we present the proposed static \name and dynamic \name. Finally, we describe how to apply \name to CNN architectures.

{\bf Overview of \name.}
Given a standard square convolutional kernel $\boldsymbol W \in \mathbb{R}^{K \times K}$ and the corresponding receptive field $ \boldsymbol{ I_{j}} \in \mathbb{R}^{K \times K}$ centered on location $j$, and letting $ \boldsymbol{\hat I_{j}} \in \mathbb{R}^{K^2 \times 1}$ and $\boldsymbol{\hat W} \in \mathbb{R}^{K^2 \times 1}$ represent the resized (from $K \times K$ to $K^2 \times 1$) RF and the kernel weight, then the corresponding output can be expressed as $\boldsymbol{O_j} = \boldsymbol{\hat W^\intercal \hat I_j}$. 

In our \name, we place a mask $\boldsymbol{M_g} \in \mathbb{R}^{K \times K}$ between the input and the kernel. The mask has a concentric distribution, which affects the kernel's perception of different regions.
The convolution can be expressed as:
\begin{equation}
\label{eq:1dGF}
\begin{aligned}
\boldsymbol{O_j} = \boldsymbol{\hat W^\intercal} (\boldsymbol{\hat M_{g} \cdot \hat I_j}) = (\boldsymbol{\hat W^\intercal \cdot \hat M_{g}^\intercal}) \boldsymbol{\hat I_j},
\end{aligned}
\end{equation}
where $\boldsymbol{\hat M_g} \in \mathbb{R}^{K^2 \times 1}$ represents the resized RF mask of  $\boldsymbol{M_g}$.
In this way, we can directly apply $\boldsymbol{M_g}$ to the kernel instead of the receptive field. RFs keep changing due to the sliding windows. So employing $\boldsymbol{M_g}$ to kernels is straightforward and efficient. 
As shown in~\cref{fig:overview_of_gmc}, the overall process can be simplified as: 
\begin{equation}
\label{eq:1dGF}
\begin{aligned}
\boldsymbol{W^{\prime}} = \boldsymbol W \otimes \boldsymbol{M_g},
\end{aligned}
\end{equation}
where $\otimes$ indicates 
Hadamard multiplication,
and $\boldsymbol{W^{\prime}} \in \mathbb{R}^{C \times K \times K}$ denotes the final RF refined kernel. The mask values are broadcasted along the channel dimension during multiplication. Details of \name are described below.





{\bf Circular Gaussian Mask.} We design the circular concentric Gaussian mask based on the one-dimensional (1D) Gaussian function:
\begin{equation}
\label{eq:1d-Gaussian func}
\begin{aligned}
f_{1\text{D}}(d) = 
\frac{1}{\sqrt{2\pi}\sigma}e^{-\frac{(d-\mu)^2}{{2\sigma}^2}},
\end{aligned}
\end{equation}
where $d$ denotes the distance between points and the center $\mu$ within the sample grids, and parameter $\sigma$ determines the distribution of the mask. Note that at $\mu$ the function takes the highest value. In particular, we put $\mu$ at the center of the convolutional kernel.

To avoid the emergence of extreme values and the absence of receptive fields, we perform $Max$ normalization over the mask:
\begin{equation}
\label{eq:1d-generate func}
\begin{aligned}
G_{1\text{D}}(d) = \frac{f_{1d}(d)}{\mathop{\max}\limits_{d \in D} f_{1d}(d)},
\end{aligned}
\end{equation}
where $D$ indicates a set of distances from the center of the kernel.
We visualize the 1D Gaussian function in \cref{fig:gauss_curve} to illustrate the potential risk if directly applying this function. 
If the $\sigma^2$ values are too large, the curve becomes very flat and all values approach to 0, causing the corresponding masked kernel to lose the overall receptive field. 
On the other hand, if the $\sigma^2$ is close to zero, the function will have very large values at the center, and the convolutional kernel can only perceive the center point. 
With the $Max$ normalization, the modified Gaussian function will always obtain the maximum value of 1 at the center, thus avoiding the emergence of extreme values and the absence of receptive fields.

{\bf Elliptical Gaussian Mask.} We design the generation function for the elliptical concentric Gaussian mask with two-dimensional Gaussian functions. The $Max$ normalization is also employed to avoid extremely large values and the absence of receptive fields:
\begin{equation}
\label{eq:2d-Gaussian func}
\begin{aligned}
f_{2\text{D}}(x,y) \!=\!
\frac{1}{2\pi\sigma_1\sigma_2} e^{-\frac{1}{2}\left(\frac{(x-\mu_1)^2}{{\sigma_1}^2}\!+\! \frac{(x-\mu_2)^2}{{\sigma_2}^2}\right)}
\end{aligned}
\end{equation}
\begin{equation}
\label{eq:2d-generate func}
\begin{aligned}
G_{2\text{D}}(x,y) = \frac{f_{2\text{D}}(x,y)}{\mathop{\max}\limits_{x,y \in K} f_{2\text{D}}(x,y)},
\end{aligned}
\end{equation}
where parameters $\sigma_1$, $\sigma_2$ determine the distribution of the mask, $(\mu_1,\mu_2)$ denotes the extreme point, $(x, y)$ denotes the offset from the center of the convolutional kernel, and $K$ represents a set of offsets within the kernel's view.



\begin{figure}[t]
  \centering
  \includegraphics[width=1\linewidth]{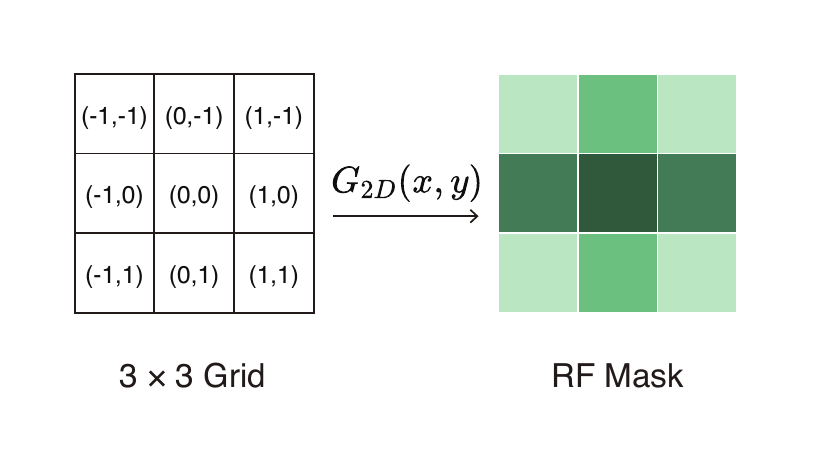}
  \caption{ Illustration of the mask generation with two-dimensional Gaussian-like function.
  }
  \label{fig:2d_mask}
\end{figure}

\begin{figure}[t]
  \centering
  \includegraphics[width=1\linewidth]{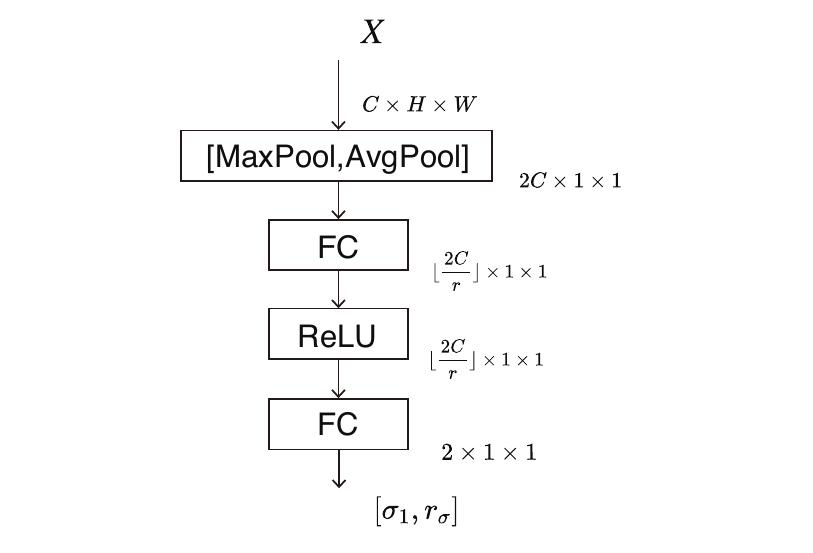}
  \caption{ Diagram of the learnable sigma module.
  }
  \label{fig:dyn_sigma}
\end{figure}

{\bf Static \name.}
We then apply the 1D generative function in~\cref{eq:1d-Gaussian func} and~\cref{eq:1d-generate func} to generate a Gaussian RF mask. Therefore, apart from learning the network weights, static \name only needs to learn one additional $\sigma$ parameter, which is initialized to 5 by default. After training, we integrate the mask into the kernel weights. As a result, it introduces NO extra inference-time computational overhead beyond the standard convolution.



{\bf Dynamic \name.}
In dynamic \name, we apply the 2D generation function to generate the RF mask as illustrated in~\cref{fig:2d_mask}.
The parameters $\sigma_1, \sigma_2$ are computed by the dynamic sigma module as shown in \cref{fig:dyn_sigma}.
Following the squeeze-and-excitation pattern, we first squeeze global information of a feature map $x\in \mathbb{R}^{C \times H \times W}$ by using both global-max-pooling and global-average-pooling operations, generating two different context vectors: $x_{max}\in \mathbb{R}^{C}$ and $x_{avg}\in \mathbb{R}^{C}$. Then we concatenate the two vectors as the global context descriptor: $z = Concat(x_{max},x_{avg})$. Next, the descriptor $z$ is forwarded to two fully connected layers (with a ReLU in-between) which output the value of $\sigma_1$ and the ratio of $\sigma_2$ to $\sigma_1$, $r_{\sigma}$. The overall process is summarized as follows:
\begin{equation}
\label{eq:learnable sigmas}
\begin{aligned}
 F_{\sigma}(z)= \boldsymbol W_1 \cdot ReLU(\boldsymbol W_0\cdot z)) + \boldsymbol B_1,
\end{aligned}
\end{equation}
where $z\in \mathbb{R}^{2C}$, $\boldsymbol W_0 \in \mathbb{R}^{\lfloor 2C/r \rfloor \times 2C}$ refers to the weights of the first fully connected layers, $\boldsymbol W_1 \in \mathbb{R}^{2 \times \lfloor 2C/r \rfloor }$ and $\boldsymbol B_1 \in \mathbb{R}^{2}$ refer to the weights and bias of the second fully connected layers, respectively, and $r$ is the reduction ratio (which is set to $4/3$ by default). 

{\bf Applying \name to CNNs}.
Our \name can be readily applied to existing CNNs and be easily trained end-to-end by standard back-propagation. When applying \name to CNNs, we replace convolutions in the stem with dynamic \name and convolutions in body with static \name.
The configurations are based on the following considerations. 
Firstly, too many dynamic \name layers can make the network difficult to train since they require joint optimization of all kernel weights and dynamic sigma modules across multiple layers.
Secondly, dynamic \name can increase the complexity and number of parameters in the network.
Finally, in CNN architecture design, the stem layer is usually kept simple, with most works focusing on the structure of the network body. Therefore, in relatively complex architectures, the convolutional kernel should be kept as simple as possible, while for simpler structures, we can enhance them with more complex convolutions.




\section{EXPERIMENTS}
\label{sec:exp}
In this section, we empirically demonstrate the effectiveness of \name on several standard benchmarks: CIFAR datasets (CIFAR10 and CIFAR100)~\cite{krizhevsky2009learning}, ImageNet~\cite{hinton2012imagenet} for image classification, and COCO 2017 for object detection~\cite{lin2014microsoft}.
We begin by comparing the performance of \name on state-of-the-art baseline models~\cite{NIPS2012_c399862d,he2016deep,sandler2018mobilenetv2, girshick2015fast,lin2017feature,cai2018cascade}. Then we conduct ablation studies to analyze different aspects of our proposed method. 
In the end, we present visualization results of \name to further analyze its effects.


\subsection{Implementation Details}
\label{sec:exp setting}
We implement our method using the publicly available PyTorch framework~\cite{paszke2017automatic}. As discussed in~\cref{sec:method}, we generally replace the first convolution with Dynamic \name (D-\name) and the remaining convolutions with Static \name (S-\name) to obtain the corresponding \name network.
Each baseline network architecture and its \name counterpart are trained from scratch with identical optimization schemes. 
The detailed settings for each dataset are as follows. 

{\bf CIFAR-10 and CIFAR-100.} For experiments on CIFAR datasets~\cite{krizhevsky2009learning}, we evaluate ResNet-20, ResNet-56, and ResNet-18~\cite{he2016deep} on both CIFAR-10 and CIFAR-100. We take random crops from the image padded by 4 pixels on each side for data augmentation and do horizontal flips. We train all the networks with an initial learning rate of 0.1, and a batch size of 128, using an SGD optimizer with a momentum of 0.9.
For ResNet-20 and ResNet-56, following the original training strategy~\cite{he2016deep}, we set their weight decay to $1e^{-4}$, and divide the learning rate at epochs 80 and 120, with a total of 160 epochs.
For ResNet-18, following Cutout~\cite{devries2017cutout}, the weight decay is set to $5e^{-4}$, and the learning rate is divided by 5 at epochs 60, 120 and 160, with a total of 200 epochs.

{\bf ImageNet.} For experiments on ImageNet, we adopt \name to AlexNet~\cite{NIPS2012_c399862d}, ResNet-18, ResNet-50~\cite{he2016deep}, and MobileNetV2~\cite{sandler2018mobilenetv2}. We use the same data augmentation scheme and training strategies as in~\cite{he2016deep,woo2018cbam} for training and apply a single-crop evaluation with the size of $224 \times 224$ at test time.
We train the above models based on the configuration from the original paper~\cite{he2016deep} or the samples from PyTorch~\cite{paszke2017automatic}.
By default, we train the networks with a batch size of $256$, using an SGD optimizer with a momentum of $0.9$ and a weight decay of $4e^{-5}$.
For MobileNetV2, the learning rate is initialized to $0.01$ and multiplied by $0.98$ for every epoch with a total of $300$ epochs. 
For AlexNet, the learning rate is initialized to $0.01$ and divided by $10$ at epochs $30$ and $60$, with a total of $90$ epochs.
For ResNets, the learning rate is initialized to $0.1$ and divided by $10$ after every $30$ epochs with a total of $90$ epochs.

{\bf COCO.} For experiments on COCO, we adopt ResNet-50 as the backbone architecture. We take the widely used Faster R-CNN architecture~\cite{renNIPS15fasterrcnn} and Cascade R-CNN~\cite{cai2018cascade} with feature pyramid networks (FPNs)~\cite{lin2017feature} as baselines, plugging in the backbones we previously trained on ImageNet.
Our training code and the parameters are based on the mmdetection toolbox~\cite{chen2019mmdetection}. The initial learning rate is set to 0.02 and we use the 1$\times$ training schedule with a batch size of 16 to train each model. Weight decay and momentum are set to 0.0001 and 0.9, respectively. We report the results using the standard COCO metrics, including $AP$ (averaged mean Average Precision over different IoU thresholds), $AP_{0.5}$, $AP_{0.75}$ and $AP_S$, $AP_M$, $AP_L$ (AP at IoU thresholds of 0.5 and 0.7, as well as AP for small, medium and large objects).

\begin{table}[!t]
\setlength\tabcolsep{2.5pt} 
\renewcommand\arraystretch{1.5}
\footnotesize
\centering
\caption{Comparisons on CIFAR datasets. We integrate the experiments of~\cref{sec:image cls} and~\cref{sec:ablation} into this table.} 
\begin{tabular}[t]{l lr rrr}
\toprule
 \multirow{2}{*}{\textbf{Model}} & \multirow{2}{*}{\textbf{Method}} & \textbf{Params} &  \textbf{FLOPs} & \textbf{CIFAR-10} & \textbf{CIFAR-100} \\
&   &   \textbf{(M)} & \textbf{(M)} & \textbf{Acc. (\%)}& \textbf{Acc. (\%)} \\

\midrule
\multirow{3}*{ResNet-20}
& StdConv & \multirow{3}*{0.27} & \multirow{3}*{42} & 91.58  & 67.03  \\
& S-\name &  &  & \bf{91.85}  & \bf{67.67}  \\
& \name &  &  & 91.81  & 67.45  \\
\cline{1-6}

\multirow{3}*{ResNet-56}
& StdConv & \multirow{3}*{0.85} & \multirow{3}*{128} & 92.86 & 69.84 \\
& S-\name &  & & 93.16   & 70.64 \\
& \name &  &  & \bf{93.28}   & \bf{70.65} \\
\cline{1-6}

\multirow{3}*{ResNet-18}
& StdConv & \multirow{3}*{11.22} & \multirow{3}*{558} & 95.14  & 77.78   \\
& S-\name &  &  & \bf{95.36}  & 78.03  \\
& \name &  &  & \bf{95.36}  & \bf{78.17}  \\

\bottomrule
\end{tabular}
\label{tab:cifar cls}
\end{table}

\subsection{Image Classification}
\label{sec:image cls}
In the following experiments, we 
denote the standard convolution as StdConv, the static \name as S-\name, and the combination of D-\name and S-\name as \name.


\begin{table}[!t]
\setlength\tabcolsep{4.5pt} 
\renewcommand\arraystretch{1}
\footnotesize
\centering
\caption{Comparisons on ImageNet. We integrate the experiments of~\cref{sec:image cls} and~\cref{sec:ablation} into this table.} 
\begin{tabular}[t]{llrrcc}
\toprule
\multirow{2}{*}{\textbf{Model}} & \multirow{2}{*}{\textbf{Method}} &  \textbf{Params} & \textbf{FLOPs} & \textbf{Top-1} & \textbf{Top-5} \\
& & \textbf{(M)} & \textbf{(G)} & \textbf{(\%)}  & \textbf{(\%)}\\
\midrule

\multirow{3}* {MobileNetV2}
& StdConv & \multirow{3}{*}{3.50} & \multirow{3}{*}{0.327} & 71.61 & 90.40\\
~& S-\name &  &  & 71.51 & 90.24 \\
~& \name &  &  & \bf{71.77} & \bf{90.29} \\
\midrule

\multirow{3}* {ResNet-18}
& StdConv & \multirow{3}{*}{11.69} & \multirow{3}{*}{1.82} & 69.77 & 89.17\\
~& S-\name &  &  & \bf{70.32} &  \bf{89.41} \\
~& \name &  &  & 69.97 & 89.22 \\
\midrule

\multirow{3}* {ResNet-50}
& StdConv & \multirow{3}{*}{25.56} & \multirow{3}{*}{4.13} & 75.55 & 92.65\\
~& S-\name &  &  & 75.85 & 92.72 \\
~& \name &  &  & \bf{76.40} & \bf{93.04} \\
\midrule

\multirow{3}* {AlexNet}
& StdConv & \multirow{3}{*}{61.10} & \multirow{3}{*}{0.714} & 56.40 & 79.24\\
~& S-\name &  &  & 57.22 & 79.63 \\
~& \name &  &  & \bf{57.38} & \bf{79.68} \\

\bottomrule
\end{tabular}
\label{tab:imagenet cls}
\end{table}

\begin{table}[!t]
\setlength\tabcolsep{3.5pt} 
\renewcommand\arraystretch{1.5}
\footnotesize
\centering
\caption{Comparisons on COCO. ResNet-50-FPN is used as the backbone.} 


\begin{tabular}[t]{l l ccc ccc}

\toprule
\multicolumn{2}{c}{\textbf{Model}}&
\textbf{AP}&\textbf{AP$_{0.5}$}&\textbf{AP$_{0.75}$}&\textbf{AP$_{S}$}&\textbf{AP$_{M}$}&\textbf{AP$_{L}$} \\
\toprule

\multirow{2}*{Faster R-CNN} & StdConv & 37.0 & 57.8 & 40.1 & 21.2 & 40.6 & 48.3 \\ 

\cline{2-8}

~& \name & \bf{37.8} & \bf{58.6} & \bf{40.8} & \bf{21.6} & \bf{41.8} & \bf{48.6} \\
\midrule

\multirow{2}*{Cascade R-CNN}
& StdConv & 40.1 & 58.4 & 43.8 & 22.3 & 43.4 & \bf{53.1} \\ \cline{2-8}
~& \name & \bf{40.6} & \bf{58.8} & \bf{44.1} & \bf{22.5} & \bf{44.2} & 53.0 \\

\bottomrule
\end{tabular}
\label{tab:coco det}
\end{table}

{\bf Results on CIFAR datasets.}
We evaluate our methods on ResNet-20, ResNet-56, and ResNet-18. 
We run each experiment three times and report the average accuracy.
The performances of each baseline and its \name counterpart on CIFAR-10 and CIFAR-100 are shown in~\cref{tab:cifar cls}.
The S-\name-related models are later discussed in~\cref{sec:ablation}.
We can observe that the amount of parameters and FLOPs for the baseline model and its \name counterpart are the same. 
This is because the extra computational cost of \name is mainly from the operation of masking filters on the stem layers, which is eliminated by integrating the mask into the kernel weights after training. Besides, the dynamic \name accounts for a small percentage of the whole network, which hardly burdens the network. Under such cases, the \name variants perform comparably to the baselines and enhance the performance of various models by a clear margin.

{\bf Results on ImageNet.} 
Table \ref{tab:imagenet cls} summarizes the results on ImageNet. Similar to the CIFAR datasets, the baseline model and its \name counterpart have the same amount of parameters and FLOPs. The Top-1 accuracy of MobileNetV2, ResNet-18, ResNet-50, and AlexNet increases by 0.16\%, 0.2\%, 0.85\% and 0.98\%, respectively. Greater gains are observed for models with greater capacity, 
especially for the large kernel network, AlexNet, that employs 11 $\times$ 11 and 5 $\times$ 5 convolutional kernels. 

\subsection{COCO Object Detection}

Table \ref{tab:coco det} compares the object detection results between our \name and standard convolutions. On the Faster R-CNN model with ResNet-50-FPN, 
we can observe that the baseline achieves an AP of 37.0\%, while applying our proposed method leads to a notable improvement of 0.8 \%.
The stronger baseline of Cascade R-CNN also benefits from incorporating \name, resulting in a clear improvement of 0.5\%.
The overall results show that \name can effectively enhance the detection performance, particularly on medium-sized objects.

\subsection{Ablation Analysis}
In this subsection, we conduct a series of ablation studies to evaluate different aspects of our proposed \name method and empirically discuss the key factors of our design.
The experimental settings are the same as in  \cref{sec:exp setting}. 

\begin{figure*}[!t]
    \centering
    \subfloat[ResNet-20]{\includegraphics[width=0.25\linewidth]{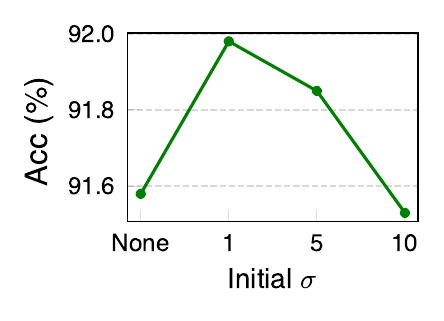}}
     \hspace{1.0em}
     \subfloat[ResNet-56]{\includegraphics[width=0.25\linewidth]{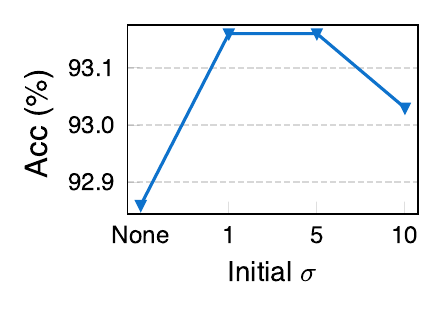}}
     \hspace{1.0em}
     \subfloat[ResNet-18]{\includegraphics[width=0.25\linewidth]{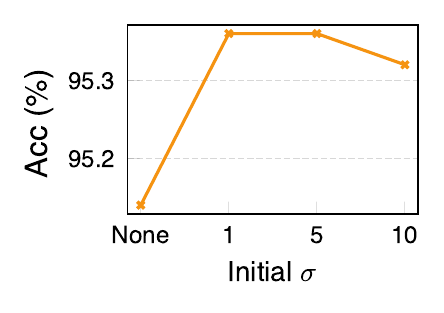}}
     \vspace{0.5em}
     \subfloat[ResNet-20]{\includegraphics[width=0.25\linewidth]{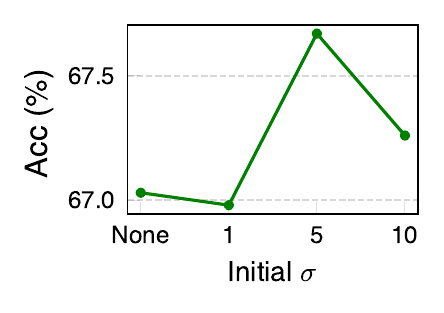}}
     \hspace{1.0em}
     \subfloat[ResNet-56]{\includegraphics[width=0.25\linewidth]{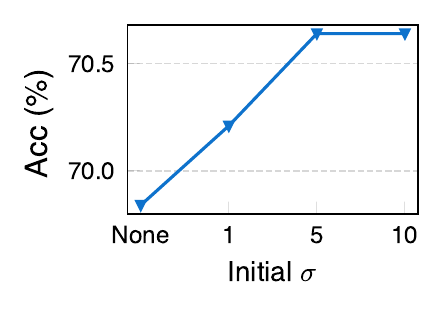}}
     \hspace{1.0em}
     \subfloat[ResNet-18]{\includegraphics[width=0.25\linewidth]{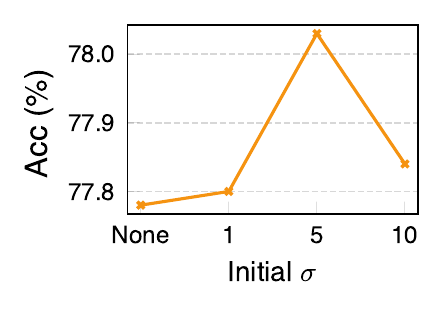}}
     
     \caption{Comparisons of S-\name with different initial $\sigma$ values on (a), (b), (c) for CIFAR-10 and (d), (e), (f) for CIFAR-100. The ``None'' term refers to the corresponding standard convolution of ConvNets. 
     }
    \label{fig:sigma_cmp_cifar100}
\end{figure*}

\label{sec:ablation}
{\bf Analysis of Static \name.} 
Tables~\ref{tab:cifar cls} and~\ref{tab:imagenet cls} also list the results of the pure static \name variants, denoted as S-\name, on CIFAR datasets and ImageNet.
The pure static S-\name variant shows promising results for most benchmark models, while allowing the convolution kernel to adjust its receptive field generally brings better performance.

However, we observe an exception in MobileNetV2.
We hypothesize that it might be because the entire MobileNetV2 adopts 3 $\times$ 3 small convolution kernels. And as we will see later on the visualization of the S-\name networks, S-\name will shrink the receptive field of the first few convolutions, whereas the receptive field of a 3 $\times$ 3 kernel is already small enough.

In addition, we find that the Top-1 accuracy of S-\name ResNet-18 is noticeably higher than the baseline and the \name variant. Such phenomenon also appears in some CNN architectures on CIFAR datasets to some extent. 
Small networks perhaps cannot effectively utilize the learnable sigma module. We did not put the results of the pure dynamic \name variant, as it is difficult to reach convergence and obtain reasonable accuracy.

\begin{table}[!t]
\renewcommand\arraystretch{1.5}
\footnotesize
\centering
\caption{Comparison of different D-\name structures on ImageNet.} 
\begin{tabular}[t]{llclc}
\toprule

\textbf{Model} & \textbf{Method}& \textbf{\#sigma} & \textbf{Pred pattern} & \textbf{Top-1(\%)}  \\
\midrule

\multirow{4}* {Resnet-18}
& StdConv & -- & --& 69.77  \\

& \name & 1 & $\sigma$ & 69.95  \\
& \name & 2 & $\sigma_1$ ,  $\sigma_2$ & 69.86  \\
& \name & 2 & $\sigma_1$ ,  $r_{\sigma}$ & \textbf{69.97}  \\

\midrule

\multirow{4}* {Resnet-50}
& StdConv & -- & --& 75.55  \\
& \name & 1 & $\sigma$ & 76.02 \\
& \name & 2 & $\sigma_1$ ,$\sigma_2 $ & 75.84 \\
& \name & 2 & $\sigma_1$, $  r_{\sigma}$ & \textbf{76.40}  \\
\midrule

\multirow{4}* {AlexNet}
& StdConv & -- & --& 56.40 \\
& \name & 1 & $\sigma$  & 57.06  \\
& \name & 2 & $\sigma_1$ ,$\sigma_2 $ & 57.07  \\
& \name & 2 & $\sigma_1$, $r_{\sigma}$ & \textbf{57.38}  \\

\bottomrule
\end{tabular}
\label{tab:GMConv_struct}
\end{table}

{\bf Analysis of the Initial $\sigma$ in S-\name.} 
As discussed in Section~\ref{sec:method}, the parameter $\sigma$ controls the receptive field of the convolutional kernels. The higher the value of $\sigma$, the larger the receptive field. Therefore, by adjusting the initial value of $\sigma$, we can analyze the effect of different initial receptive fields.
We compare three initial values, 1, 5 and 10, on CIFAR-10 and CIFAR-100. Specifically, we train ResNet-20, ResNet-56, ResNet-18 and their S-\name variants three times and report the average accuracy. 

\cref{fig:sigma_cmp_cifar100} provides the comparisons on CIFAR-10 and CIFAR-100. On CIFAR-10, the networks tend to perform better with a small initial receptive field, while on CIFAR-100, either too small or too large initial receptive fields may compromise the performance of the networks considerably.
It demonstrates the importance of a suitable receptive field for models.
A proper initial receptive field can further improve the performance, whereas too small or too large receptive fields may harm the models. The results also suggest that 5 is an appropriate initial receptive field value, which can steadily improve the model performance. 

{\bf Designs of D-\name.}
We design our learnable sigma module to predict the mask generation parameters. 
When the module predicts only one value, that value represents the magnitude of $\sigma$, denoted as $\sigma$. If the network predicts two values, they may be the magnitudes of $\sigma_1$ and $\sigma_2$, denoted as $(\sigma_1,\sigma_2)$, or the magnitude of $\sigma_1$ and the scale factor $r_{\sigma}$, denoted as $(\sigma_1,r_{\sigma})$.

As shown in Table~\ref{tab:GMConv_struct}, generally, if the models simply predict the magnitude of sigma values (denoted as $\sigma$ and $\sigma_1$, $\sigma_2$), increasing the number of sigma parameters does not lead to better performance despite a more flexible RF mask.
However, it does not imply that the elliptical RF mask is inferior to the circular RF mask.
We found that the performance of the two-parameter ($\sigma_1$, $\sigma_2$) pattern can be improved by predicting $\sigma_1$ and the ratio ($r_{\sigma}$) between the two sigmas, where $\sigma_2$ can be calculated by $\sigma_1 \times r_{\sigma}$. 

The above experiments demonstrate the effectiveness of our D-\name. 
On the other hand, it also shows that a simple modification to the stem convolution can considerably impact the overall network. Despite the small proportion of the stem layer in CNNs, 
our results suggest that careful design of this layer remains necessary for optimal performance.

\begin{figure}[tb]
  \centering
  \includegraphics[width=\linewidth]{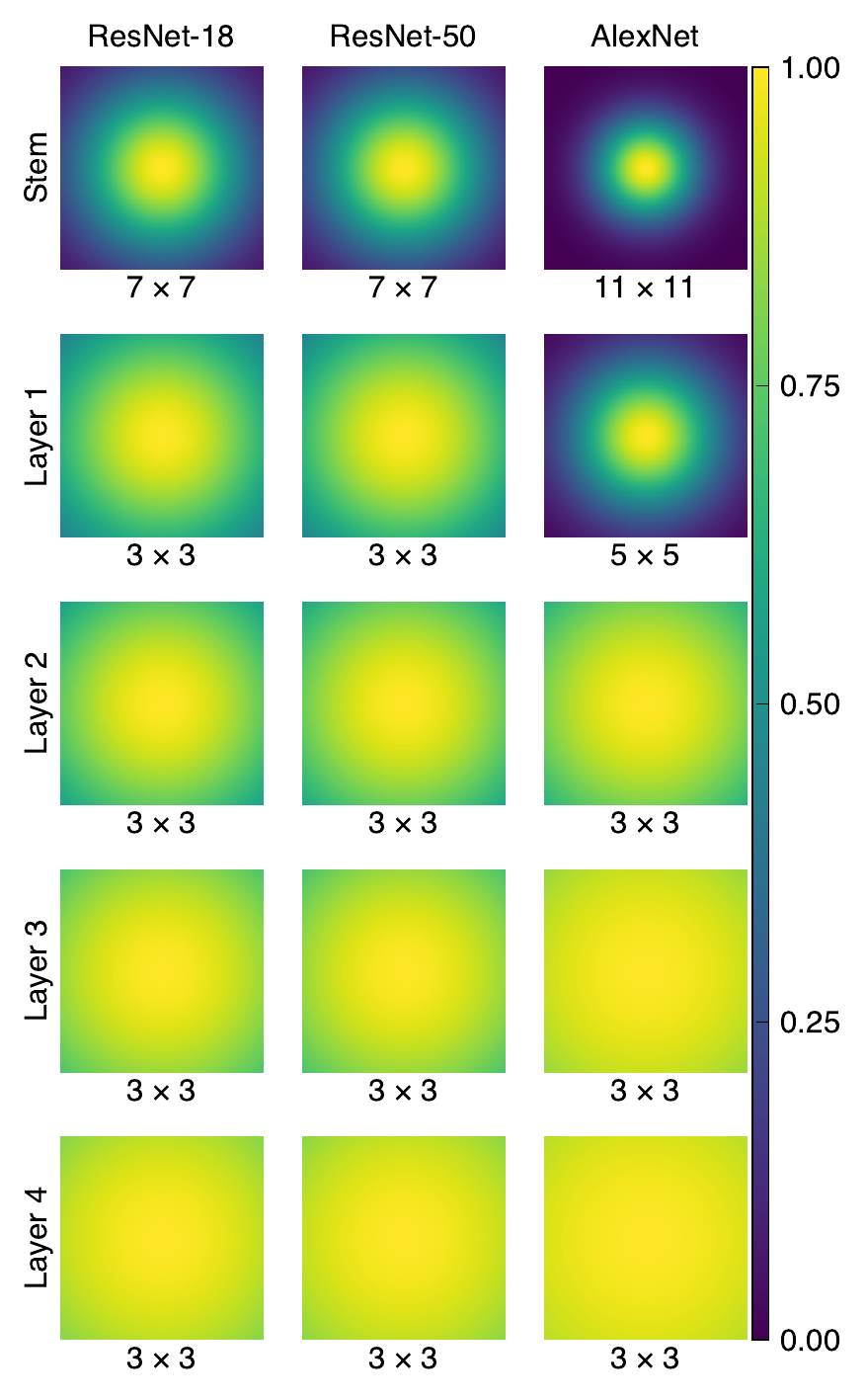}
  \caption{RF masks visualization of S-\name networks. We visualize the RF masks of different layer convolutions in ResNet-18, ResNet-50 and AlexNet. In particular, we calculate the average RF masks of ResNet-18, ResNet-50, AlexNet in different body layers.
  }
  \label{fig:RF_visual}
\end{figure}

\subsection{Visualization}

{\bf RF Mask Visualization.}
To investigate the effect of RF masks on the network, we visualize in~\cref{fig:RF_visual} the RF masks of different layers for S-\name CNNs. It can be observed that the S-\name generally affects the receptive field of the shallow layers of CNN, especially for large convolutional kernels, while in deeper layers, the RFs are not much different from the standard convolution.


It shows that, 
in shallow layers of the network, smaller receptive fields (RF) are more favorable, while in deeper layers, larger RFs are preferred. We then conduct several experiments on COCO object detection where \name is applied to the first few convolutional layers, and we combine \name with deformable convolution~\cite{dai2017deformable} which are generally used in deep layers. When combining \name with deformable convolution, we replace the convolutions in the stem and the first body layer with \name, while the rest of the convolutions in backbone remain as deformable convolutions by default.

The comparison results are presented in Table~\ref{tab:usage of gmc}. It is worth noting that, in general, using \name before the third body layer in ResNet is a preferable choice, whereas using it in deeper layers may compromise the model's performance.
Additionally, the utilization of \name is shown to enhance the detection of small and medium objects. To a certain extent, increasing the number of network layers that incorporate \name can boost the performance on small object detection.
Finally, while \name performance falls short to deformable convolution, combining the two can yield a performance improvement for both detectors.

Our results suggest that the advantages of GMConv stem from its ability to use smaller kernels in the early layers. 
By using a Gaussian mask to adjust the effective kernel size, \name allows for continuous adjustment of the kernel size with minimal overhead.

\begin{figure*}[tb]
  \centering
  \includegraphics[width=1\linewidth]{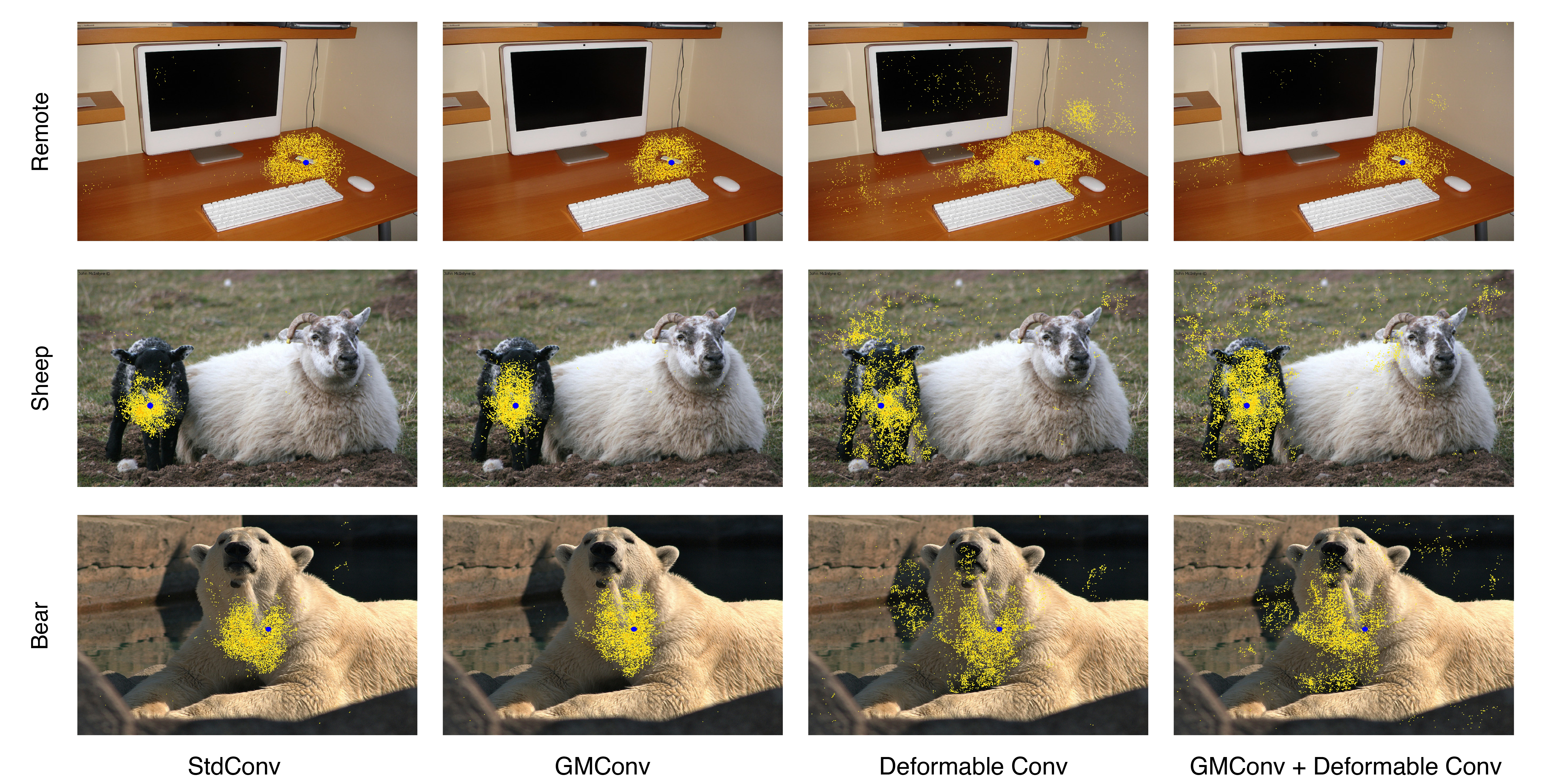}
  \caption{Visualization of Effective Receptive Fields (ERFs). 
  We compare the ERFs of standard convolution, \name, deformable convolution, and combination of \name and deformable convolution. The baseline is Faster R-CNN + ResNet-50-FPN.
  }
  \label{fig:erf_vis}
\end{figure*}

{\bf Effective Receptive Fields Visualization.}
The above experiments demonstrate that using \name in early layers results in greater precision in object localization and potentially more detailed low-level features. We further visualize the effective receptive fields learned by different convolutional kernels for objects of varying sizes in object detection. 
As shown in~\cref{fig:erf_vis}, our results demonstrate that \name exhibits a more compact effective receptive field than standard convolution when detecting small objects, while it has a more expansive effective receptive field for larger objects. In contrast, the effective receptive field of deformable convolution tends to produce more dispersed effective receptive fields that cover larger areas. 
We also show that combining with \name can effectively alleviate the dispersion of deformable convolution and obtain a more precise and dense effective receptive field.

\begin{table*}[!t]
\renewcommand\arraystretch{1.5}
\footnotesize
\centering
\caption{Results of using \name in stem and the first 1, 2, 3 and 4 body layers in ResNet-50, and results of \name and deformable convolution.} 

\begin{tabular}[t]{ll cccc c cccc}
\toprule
\multirow{2}*{\textbf{Method}}&
 \textbf{Usage of}&\multicolumn{4}{c}{\textbf{Faster R-CNN}} & &\multicolumn{4}{c}{\textbf{Cascade R-CNN}}\\
 \cline{3-6}  \cline{8-11}

 ~& \textbf{\name} &
AP &AP$_{S}$&AP$_{M}$&AP$_{L}$ & &
AP &AP$_{S}$&AP$_{M}$&AP$_{L}$\\
\toprule

\multirow{5}*{\name}
& None (Baseline) & 37.0 & 21.2 & 40.6 & 48.3 & ~ 
& 40.1 & 22.3 & 43.4 & \bf{53.1}\\
~ & Stem L1 & \bf{37.9}  & 21.7  & \bf{42.0}  & \bf{48.7} & ~
 & 40.6  & 22.5  & 44.2  & 53.0   \\
~& Stem L1-L2 & \bf{37.9} & 22.1 & 41.9 & 48.4 & ~ &
\bf{40.7} & \bf{23.0} & 44.3 & 52.7 \\
~& Stem L1-L3 & 37.8 & \bf{22.2} & 41.7 & 48.5 & ~ &
40.6 & 22.5 & \bf{44.5} & 52.7 \\
~& Stem L1-L4 &37.8 & 21.6 & 41.8 & 48.6 & ~ &
40.6 & 22.5 & 44.2 & 53.0\\
\midrule

DCN (L2-L4) & None & 41.1 & 23.6 & 44.7 & \bf{54.9} & ~ &
43.3 & 24.3 & 47.0 & 57.8 \\ 
\name-DCN & Stem L1 & \bf{41.5} & \bf{24.4} & \bf{45.4} & 54.4 & ~ 
&\bf{43.8} & \bf{25.8} & \bf{47.5} & \bf{58.3} \\

\bottomrule
\end{tabular}
\label{tab:usage of gmc}
\end{table*}

\section{Conclusion}
\label{sec:con}
Inspired by the Gaussian distribution of effective receptive fields in CNNs, we propose the Gaussian Mask Convolutional Kernel (\name) that introduces a concentric receptive field to convolutional kernels. 
Existing related works mainly focus on sampling pixels with offsets, which cannot model the effective receptive field effectively. 
In contrast, \name modifies the Gaussian function to generate receptive field masks which then are put over the convolutional kernels to adjust receptive fields within the kernel. In particular, we provide a static version and a dynamic version of \name that work on different layers of the network to achieve a good trade-off between effectiveness and complexity.
Our \name can be easily integrated into existing CNN architectures. Experiments on image classification and object detection demonstrate that \name can substantially boost the performance of equivalent networks.
Our visualization results indicate that \name mainly takes effect in shallower layers, where smaller receptive fields are preferred, and larger receptive fields are preferred in deeper layers.
We hope that our findings will provide insights for 
better designs 
 of convolutional neural networks.


\section*{Acknowledgments}
This work is supported by National Natural Science Foundation (U22B2017,62076105). 

\bibliographystyle{IEEEtran}
\bibliography{IEEEabrv,egbib}

\end{document}